\documentclass[11pt,a4paper]{article}
\usepackage[hyperref]{eacl2021}
\usepackage{times}
\usepackage{latexsym}

\usepackage{microtype}
\usepackage{hyperref}
\usepackage{tabularx}
\usepackage{graphicx}
\usepackage{multirow}

\aclfinalcopy 

\title{\textsc{AraELECTRA}: Pre-Training Text Discriminators\\ for Arabic Language Understanding}

\author{Wissam Antoun \and Fady Baly \and Hazem Hajj \\
        American University of Beirut\\
        \{wfa07, fbg06, hh63\}@aub.edu.lb}
        
\date{}

\begin{document}
\maketitle
\begin{abstract}
Advances in English language representation enabled a more sample-efficient pre-training task by Efficiently Learning an Encoder that Classifies Token Replacements Accurately (ELECTRA).
Which, instead of training a model to recover masked tokens, it trains a discriminator model to distinguish true input tokens from corrupted tokens that were replaced by a generator network. 
On the other hand, current Arabic language representation approaches rely only on pretraining via masked language modeling.
In this paper, we develop an Arabic language representation model, which we name \textsc{AraELECTRA}.
Our model is pretrained using the replaced token detection objective on large Arabic text corpora.
We evaluate our model on multiple Arabic NLP tasks, including reading comprehension, sentiment analysis, and named-entity recognition and we show that \textsc{AraELECTRA} outperforms current state-of-the-art Arabic language representation models, given the same pretraining data and with even a smaller model size.
\end{abstract}

\section{Introduction}
Recently, pre-trained language representation models have demonstrated state-of-the-art performance on multiple NLP tasks and in different languages.
Pre-training is commonly done via Masked Language Modeling (MLM)~\cite{devlin2019bert, liu2019roberta, conneau2019unsupervised}, where an input sequence has some of its tokens randomly hidden and the model is tasked to recover the original masked tokens.
While this approach has proven successful, recent works have shown that MLM is not sample-efficient~\cite{clark2020electra}, since the network only learns from the small subset of masked tokens per sequence (15\% of the tokens in BERT).
\citet{clark2020electra} proposed an approach called  Efficiently Learning an Encoder that Classifies Token Replacements Accurately (ELECTRA).
The method uses a pre-training technique based on replaced token detection (RTD) task is more efficient than MLM, and thus achieved state-of-the-art results on English benchmarks.
RTD is a pre-training task where a model is tasked to distinguish true input tokens from synthetically generated ones.
RTD solves the issue of the mismatch created in MLM, where the model only sees the \texttt{[MASK]} token during pre-training but not during fine-tuning.
In ELECTRA, a small masked language generator network \textbf{G} is used to generate used to generate the corrupted tokens, and BERT-based discriminator model \textbf{D} predicts for whether a token is an original or a replacement.

Current state-of-the-art language representation models for Arabic employ MLM as a pre-training objective~\cite{antoun2020arabert,safaya-etal-2020-kuisail,lan2020gigabert,abdul2020toward,chowdhury-etal-2020-improving-arabic,abdul2020arbert}.
In this paper, we describe the process of pre-training a transformer encoder model for Arabic language understanding using the RTD objective, which we call \textsc{AraELECTRA}.
We also evaluate \textsc{AraELECTRA} on multiple Arabic NLP tasks and show empirically that \textsc{AraELECTRA} outperforms current state-of-the-art Arabic pre-trained models.

Our contributions can be summarized as follows:
\begin{itemize}
\setlength\itemsep{-0.3em}
    \item Pre-training the ELECTRA model on a large-scale Arabic corpus.
    \item Reaching a new state-of-the-art on multiple Arabic NLP tasks.
    \item Publicly releasing \textsc{AraELECTRA} on popular NLP libraries.
\end{itemize}

The rest of the paper is organized as follows. 
Section~\ref{sec:rel} provides a review of previous Arabic language representation literature.
Section~\ref{sec:method} details the methodology used in developing \textsc{AraELECTRA}. 
Section~\ref{sec:eval} describes the experimental setup, evaluation procedures, and experiment results.
Finally, we conclude in Section~\ref{sec:conc}.

\begin{figure*}[!ht]
  \includegraphics[width=\textwidth]{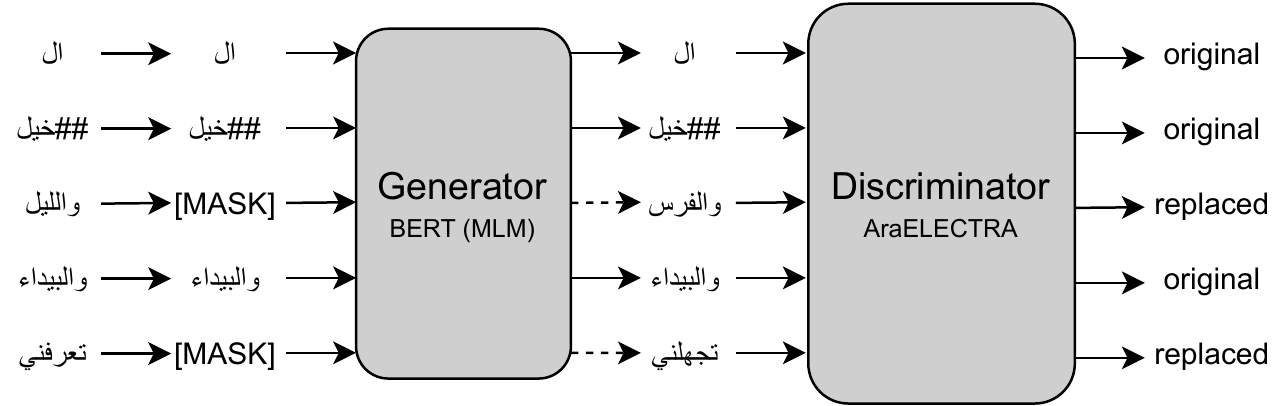}
  \caption{Replaced Token Detection pre-training task\label{electrafig}}
\end{figure*}
\section{Related Works}
\label{sec:rel}
Recently, work on Arabic language representation have been on the rise due to the performance benefits that transfer learning approaches have brought.
Early transfer learning approaches in Arabic relied on using pre-trained word embeddings i.e. AraVec~\cite{soliman2017aravec}.
Model-level transfer learning was shown to work on Arabic with hULMonA~\cite{eljundi2019hulmona}, a recurrent neural network-based language modeling approach.
\citet{antoun2020arabert} and \citet{safaya-etal-2020-kuisail} improved on hULMonA, and pre-trained transformer-based models with MLM with large scale Arabic corpora.
Other approaches addressed issues with the early BERT-based models such as training on code-switched English-Arabic corpora to improve performance on information retrieval tasks~\cite{lan2020gigabert}, and training on dialectal Arabic (DA) corpora to address the domain miss-match between MSA and DA during pre-training and finetuning~\cite{abdul2020toward,chowdhury-etal-2020-improving-arabic}.

We hence propose an Arabic ELECTRA-based language representation model pre-trained using the RTD objective on large MSA corpora.

\section{\textsc{AraELECTRA}: Methodology}
\label{sec:method}
In this paper, we develop an ELECTRA-based Arabic language representation model to improve the state-of-the-art in Arabic reading comprehension.
We create \textsc{AraELECTRA} a bidirectional transformer encoder model with 12 encoder layers, 12 attention heads, 768 hidden size, and 512 maximum input sequence length for a total of 136M parameters.
The pre-training setup and dataset of \textsc{AraELECTRA} are described in the following sections.

\subsection{Pre-training Setup}
While \textsc{AraBERT} was trained using the MLM objective, \textsc{AraELECTRA} is pre-trained using the RTD objective.
The RTD approach trains two neural network models, a generator \textbf{G} and a discriminator \textbf{D} or \textsc{AraELECTRA}, as shown in Figure~\ref{electrafig}.
\textbf{G} takes a corrupted input sequence, where random tokens are replaced with the \texttt{[MASK]} token, and learns to predict the original tokens that have been masked.
The generator network \textbf{G} is in our case a small BERT model with 12 encoder layers, 4 attention heads, and 256 hidden size\footnote{In the generator, the input embeddings of size 768 are first projected into the generator hidden size with the addition of a linear layer.}. 
The discriminator network \textbf{D} then takes as input the recovered sequence from the output of \textbf{G} and tries to predict which tokens were replaced and which tokens are from the original text.

While this approach may look similar to a generative adversarial network (GAN)~\cite{goodfellow2014generative}, the generator network in \textsc{ELECTRA} is trained with maximum-likelihood instead of adversarial training to fool the discriminator and the input to the generator is not a random noise vector, but a corrupted sequence of tokens.

\subsection{Pre-training Dataset}
We chose to pre-train on the same dataset as \textsc{AraBERTv0.2}~\cite{antoun2020arabert}, to make the comparison between models fair.
The dataset is a collection of the Arabic corpora list below:
\begin{itemize}
\setlength\itemsep{-0.3em}
    \item The OSCAR corpus~\cite{ortiz-suarez-etal-2020-monolingual}.
    \item The 1.5B words Arabic Corpus~\cite{el20161}.
    \item The Arabic Wikipedia dump from September 2020.
    \item The OSIAN corpus~\cite{zeroual2019osian}.
    \item News articles provided by As-Safir newspaper.
\end{itemize}
The total size of the training dataset is 77GB or 8.8 billion words, and comprises mostly news articles.
For validation, we use new Wikipedia articles that were published after the September 2020 dump.

The same wordpiece vocabulary from \textsc{AraBERT}v0.2 was used for tokenization.

\subsection{Fine-tuning}
Since the discriminator network has the same architecture and layers as a \textsc{BERT} model, we add a linear classification layer on top of \textsc{ELECTRA}'s output, and fine-tune the whole model with the added layer on new tasks.
\textsc{AraELECTRA}'s performance is validated on three Arabic NLP tasks i.e. question answering (QA), sentiment analysis (SA) and named-entity recognition (NER).

\section{Experiments and Evaluation}
\label{sec:eval}
\subsection{Experimental Setup}
\paragraph{pre-training}
For pre-training, 15\% of the 512 input tokens were masked.
The model was pre-trained for 2 million steps with a batch size of 256.
Pre-training took 24 days to finish on a TPUv3-8 slice.
The learning rate was set to 2e-4, with 10000 warm-up steps.

\paragraph{Fine-tuning}
All the models were fine-tuned with batch size set to 32, maximum sequence length of 384, and a stride of 128 for QA, and a maximum sequence length of 256 for SA and NER. 
Experiments were only performed with the following learning rates [2e-5, 3e-5, 5e-5], since model specific hyper-parameter optimization is computationally expensive.

\subsection{Datasets and other Models}
\subsubsection{Question Answering}
The question answering task examines the model's reading comprehension and language understanding capabilities.
The datasets of choice are the Arabic Reading Comprehension Dataset (ARCD)~\cite{mozannar-etal-2019-neural} and the Typologically Diverse Question Answering dataset (TyDiQA)~\cite{tydiqa}.
Both datasets follow the SQuAD~\cite{rajpurkar2016squad} format where the model is required to extract the span of the answer, given a question and a context.

The ARCD \cite{mozannar-etal-2019-neural} training set consists of 48344 machine-translated questions and answers from English, with 693 questions and answers from the ARCD set.
The test was performed on the remaining 702 questions from the ARCD set.
From the TyDiQA~\cite{tydiqa}, we chose the Arabic examples from the training and development sets of subtask 2, for a total of 14508 pairs for training and 921 pairs for testing.

\subsubsection{Sentiment Analysis}
Arabic sentiment Analysis evaluation is done on the Arabic Sentiment Twitter Dataset for LEVantine (ArSenTD-Lev)~\cite{baly2018arsentd}.
The dataset contains 4000 tweets written in the Levantine Arabic dialect and annotated for the sentiment (5 classes), topic, and sentiment target.
The data was split 80-20 for training and testing.

\subsubsection{Named-Entity Recognition}
For Arabic NER recognition, the model is evaluated on the ANERcorp dataset~\cite{benajiba2007}, with the data split from CAMeL Lab~\cite{obeid-etal-2020-camel}. 
The train split has 125,102 words and the test split has 25,008 words, labeled for organization (ORG), person (PER), location (LOC), and miscellaneous (MISC).

\subsubsection{Reference Models}
We evaluate our model against a collection of Arabic transformer models.
\begin{itemize}
\setlength\itemsep{-0.3em}
    \item \textsc{AraBERTv0.1}~\cite{antoun2020arabert}.
    \item \textsc{AraBERT}v0.2 base, large~\cite{antoun2020arabert}.
    \item \textsc{Arabic-BERT} base, medium, large~\cite{safaya-etal-2020-kuisail}.
    \item \textsc{Arabic ALBERT} base, large, xlarge\footnote{https://github.com/KUIS-AI-Lab/Arabic-ALBERT/}.
    \item \textsc{ARBERT}~\cite{abdul2020arbert}.
\end{itemize}

\begin{table*}[ht]
\centering
\resizebox{0.9\textwidth}{!}{%
\begin{tabular}{lcccccc}
\hline
\multicolumn{1}{c}{\multirow{2}{*}{Model}} & \multicolumn{2}{c}{TyDiQA} & \multicolumn{2}{c}{ARCD} & \multicolumn{1}{c}{ArSenTD-LEV} & \multicolumn{1}{c}{ANERcorp} \\
\multicolumn{1}{c}{} & \multicolumn{1}{c}{EM} & \multicolumn{1}{c}{F1} & \multicolumn{1}{c}{EM} & \multicolumn{1}{c}{F1} & \multicolumn{1}{c}{F1} & \multicolumn{1}{c}{F1} \\ \hline
AraBERTv0.1 & 68.51 & 82.86 & 31.62 & 67.45 & 53.56 & 83.14\\
AraBERTv0.2-base & 73.07 & 85.41 & 32.76 & 66.53 & 55.71 & 83.70 \\
AraBERTv0.2-large & 73.72 & 86.03 & 36.89 & \textbf{71.32} & 56.94 & 83.08 \\ \hline
Arabic-BERT-base & 67.42 & 81.24 & 30.48 & 62.24 & 54.21 & 81.05 \\
Arabic-BERT-large & 70.03 & 84.12 & 33.33 & 67.28 & 55.32 & 82.15 \\ \hline
Arabic-ALBERT-base & 67.10 & 80.98 & 30.91 & 61.33 & 51.70 & 76.89 \\
Arabic-ALBERT-large & 68.07 & 81.59 & 34.19 & 65.41 & 54.62 & 79.61 \\
Arabic-ALBERT-xlarge & 71.12 & 84.59 & \textbf{37.75} & 68.03 & 54.15 & 81.13 \\ \hline
ARBERT & 71.55 & 83.69 & 31.62 & 65.93 & 53.52 & 83.33 \\ \hline
AraELECTRA & \underline{\textbf{74.91}} & \underline{\textbf{86.68}} & \underline{37.03} & \underline{71.22} & \underline{\textbf{57.20}} & \underline{\textbf{83.95}}
\end{tabular}%
}
\caption{Performance of all tested model on the various Arabic downstream tasks. Overall best scores are highlighted in bold, while the best score within base-sized models is underlined\label{results}.}
\end{table*}

\subsection{Results}
Experimental results for the different datasets and models are shown in Table~\ref{results}.

The results show that \textsc{AraELECTRA} achieved the highest performance on all tested datasets when compared to the other base models, and only fell short on ARCD to Arabic-\textsc{ALBERT}-xlarge, a model 4 times its size, in exact match scores, and to \textsc{AraBERT}v0.2-large in F1-score.

The performance difference between both QA datasets is due to the poor quality of the ARCD training examples, which are translated from English SQuAD.
ARCD training examples also contained text in languages other than Arabic and English, which further reduced performance due to the occurrence of unknowns subwords and characters.
It is also to be noted, that some training examples in Arabic TyDiQA contained HTML artifacts which appeared in the training context and answer.

As for the ArSenTD-LEV scores, all test Arabic models still struggle with fine-grained labelling of ArSenTD-Lev.
Mainly because the dataset only contains 4K examples distributed between 5 sentiment classes and on 6 diverse topics, with high class-imbalance.

These results clearly demonstrate that \textsc{ELECTRA}'s RTD objective achieves higher performance especially on QA tasks and improved semantic representation compared to MLM on Arabic text.
\section{Conclusion}
\label{sec:conc}
In this paper, we showed that pre-training using the RTD objective on Arabic text is more efficient and produces pre-trained language representation models better than the MLM objective.
Our \textsc{AraELECTRA} model improves the state-of-the-art for Arabic Question Answering, sentiment analysis and named-entity recognition, and achieves higher performance compared to other models pre-trained with the same dataset and with larger model sizes.
Our model will be publicly available, along with our pre-training and fine-tuning code, in our repository \href{https://github.com/aub-mind/arabert/tree/master/araelectra}{github.com/aub-mind/arabert/tree/master/araelectra}

\section*{Acknowledgments}
The author would like to thank Tarek Naous for the constructive criticism of the manuscript.
This research was supported by the University Research Board (URB) at the American University of Beirut (AUB), and by the TFRC program, which we thank for the free access to cloud TPUs.
We also thank As-Safir newspaper for the data access.

\bibliography{references.bib}
\bibliographystyle{acl_natbib}

\end{document}